\documentclass[11pt,a4paper]{article}
\usepackage[hyperref]{acl2021}
\usepackage{times}
\usepackage{latexsym}

\usepackage{microtype}
\aclfinalcopy % Uncomment this line for the final submission

\usepackage[utf8]{inputenc}
\usepackage{graphicx}
\usepackage{url}
\usepackage{booktabs}
\usepackage{bbm}
\usepackage{amsmath}
\usepackage{amsthm}
\usepackage{amssymb}
\usepackage{algorithm}
\usepackage{algorithmic}
\usepackage{subfigure}
\usepackage{cases}
\usepackage{multirow}
\usepackage{multicol}

\newcommand{\mymethod}[0]{\textsc{ReRe} }

\newcommand{\bci}[0]{\mathbf{c}_i}
\newcommand{\sro}[0]{\langle s,r,o \rangle}
\newcommand{\hyrc}[0]{\mathbf{\hat{y}}_{rc}}

\newcommand{\hy}[0]{\hat{y}}
\newcommand{\by}[0]{\mathbf{y}}
\newcommand{\hby}[0]{\mathbf{\hat{y}}}
\newcommand{\hyee}[0]{\mathbf{\hat{y}}_{ee}}

\newcommand{\bw}[1]{\mathbf{W}_{#1}}
\newcommand{\bb}[1]{\mathbf{b}_{#1}}
\newcommand{\ep}[0]{\mathbb{E}}

\title{Revisiting the Negative Data of Distantly Supervised Relation Extraction}

\author{Chenhao Xie$^{1,2}$, Jiaqing Liang$^1$, Jingping Liu$^1$, Chengsong Huang$^1$, Wenhao Huang$^1$, Yanghua Xiao$^1$\\
$^1$Fudan University, Shanghai, China\\
\{redreamality, l.j.q.light\}@gmail.com \quad \{jpliu17,huangcs19,whhuang17,shawyh\}@fudan.edu.cn
  }

\date{}

\begin{document}
\maketitle
\begin{abstract}
Distantly supervision automatically generates plenty of training samples for relation extraction. 
However, it also incurs two major problems: noisy labels and imbalanced training data.
Previous works focus more on reducing wrongly labeled relations (false positives) while few explore the missing relations that are caused by incompleteness of knowledge base (false negatives).
Furthermore, the quantity of negative labels overwhelmingly surpasses the positive ones in previous problem formulations.
In this paper, we first provide a thorough analysis of the above challenges caused by negative data.
Next, we formulate the problem of relation extraction into as a positive unlabeled learning task to alleviate false negative problem. 
Thirdly, we propose a pipeline approach, dubbed \textsc{ReRe}, that performs sentence-level relation detection then subject/object extraction to achieve sample-efficient training.
Experimental results show that the proposed method consistently outperforms existing approaches and remains excellent performance even learned with a large quantity of false positive samples.
\end{abstract}

\section{Introduction}
\begin{figure}[t]
  \centering
  \includegraphics[width=\columnwidth]{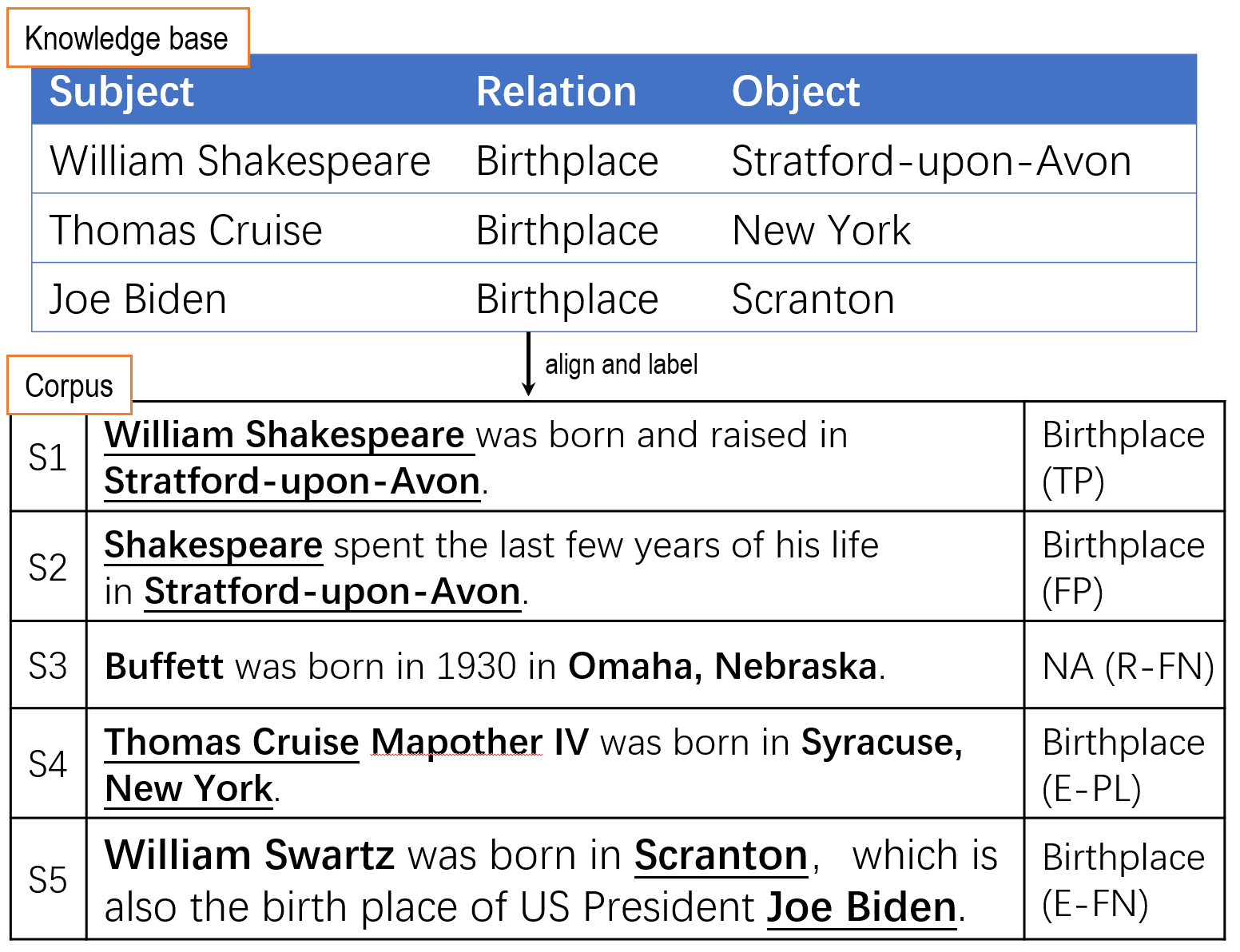}
  \caption{Illustration of distant supervision process. S2-S5 are examples for four kinds of label noise. 
  TP, FP, FN and PL mean \textit{true positive, false positive, false negative} and \textit{partially labeled}, respectively.
  ``R-" or ``E-" indicates whether the error occurs at \textit{relation-level} or \textit{entity-level}. \textbf{Bold tokens} are ground-truth subjects/objects. \underline{Underlined tokens} together with the relation in the third column are labeled by distant supervision. ``NA" means no relation.}
  \label{fig:re_example}
\end{figure}

Relational extraction is a crucial step towards knowledge graph construction.
It aims at identifying relational triples from a given sentence in the form of $\langle$subject, relation, object$\rangle$, in short, $\langle s,r,o \rangle$.
For example, given S1 in Figure \ref{fig:re_example}, we hope to extract $\langle$\textsc{William Shakespeare, Birthplace, stratford-upon-Avon}$\rangle$.

% Jingping Liu

This task is usually modeled as a supervised learning problem and \textit{distant supervision} \cite{mintz2009distant} is utilized to acquire large-scale training data. 
The core idea is to obtain training data is through automatically labeling a sentence with existing relational triples from a knowledge base (KB). 
For example, given a triple $\langle s,r,o \rangle$ and a sentence, if the sentence contains both $s$ and $o$, distant supervision methods regard $\sro$ as a valid sample for the sentence.
If no relational triples are applicable, the sentence is labeled as ``NA". 
% If a sentence matches the subjects and objects , those relations will also be marked as labels for the sentence. 

Despite the abundant training data obtained with distant supervision, nonnegligible errors also occur in the labels.
There are two types of errors. 
In the first type, the labeled relation does not conform with the original meaning of sentence, and this type of error is referred to as \textit{false positive} (FP). 
For example, in \textit{S2}, The ``\textit{Shakespeare spent the last few years of his life in Stratford-upon-Avon.}" does not express the relation \textsc{Birthplace}, thus being a FP. 
Second, large amounts of relations in sentences are missing due to the incompleteness of KB, which is referred to as \textit{false negative} (FN). 
For instance, in \textit{S3}, ``\textit{Buffett was born in 1930 in Omaha, Nebraska.}" is wrongly labeled as NA since there is no relation (e.g., \textsc{Birthplace}) between \textsc{Buffett} and \textsc{Omaha, Nebraska} in the KB. 
Many efforts have been devoted to solving the FP problem, including pattern-based methods~\cite{jia2019arnor}, multi-instance learning methods~\cite{lin2016neural,zeng2018large} and reinforcement learning methods~\cite{feng2018reinforcement}. Significant improvements have been made. 

\textit{However, FN problem receives much less attention}~\cite{min2013distant,xu2013filling,roller2015improving}.
To the best of our knowledge, none existing work with deep neural networks to solve this problem.
We argue that this problem is fatal in practice since there are massive FN cases in datasets. 
For example, there exist at least 33\% and 35\% FN's in NYT and SKE datasets, respectively. We will deeply analyze the problem in Section~\ref{sec:fn}

Another huge problem for relation extraction is \textit{overwhelming negative labels}.
%negative data is \textit{imbalanced class distribution}.
As is widely acknowledged, information extraction tasks are highly imbalanced in class labels~\cite{Chowdhury2012ImpactOL,lin2018adaptive,Li2020DiceLF}.
In particular, the negative labels account for most of the labels in relation extraction under almost any problem formulation, which makes relation extraction a hard machine learning problem. 
We systematically analyze this in Section~\ref{sec:neglabel}.

In this paper, we address these challenges caused by negative data. Our main contribution can be summarized as follows. 
\begin{itemize}
    \item We systematically compare the class distributions of different problem modeling and explain why first extract relation then entities, i.e., the third paradigm (P3) in Section~\ref{sec:neglabel},  is superior to the others.
    \item Based on the first point, we adopt P3 and propose a novel two-staged pipeline model dubbed \textsc{ReRe}. 
    It first detects relation at sentence level and then extracts entities for a specific relation.
    We model the false negatives in relation extraction as ``unlabeled positives" and propose a multi-label collective loss function.
    \item Our empirical evaluations show that the proposed method consistently outperforms existing approaches, and achieves excellent performance even learned with a large quantity of false positive samples. 
    We also provide two carefully annotated test sets aiming at reducing the false negatives of previous annotation, namely, NYT21 and SKE21, with 370 and 1150 samples, respectively.
\end{itemize}

\section{Problem Analysis and Pilot Experiments}

We use $(\bci, T_i)$ to denote a training instance, where $\bci$ is a sentence consisting of $N$ tokens $\bci=[c_{i1},...,c_{iN}]$ labeled by a set of triples $T_i = \{\sro\}$ from the training set $\mathcal{D}$.
For rigorous definition, $[c_{i1},...,c_{iN}]$ can be viewed as an ordered set $\{(c_{i1},1), ..., (c_{iN},N)\}$ so that set operations can be applied.
% $s$ and $o$ are 
We assume $r \in \mathcal{R}$, where $\mathcal{R}$ is a finite set of all relations in $\mathcal{D}$. 
% $\yrc \in \{0,1\}^{|\mathcal{R}|}$ and $\yee \in \{0,1\}^{N \times 4}$ are the labels for \textit{relation classification} and \textit{entity extraction} model, respectively (the models are later defined in Section \ref{sec:model}).
Other model/task-specific notations are defined after each problem formulation.

% TODO check all the usage of following terms
We now clarify some terms used in the introduction and title without formal definition.
A \textbf{negative sample} refers to a triple  $t \notin T_i$.
\textbf{Negative label} refers to the negative class label (e.g., usually ``0'' for binary classification), used for supervision with respect to task-specific models. Under different task formulation, the negative labels can be different.
\textbf{Negative data} is a general term that includes both negative labels and negative samples.
There are two kinds of \textit{false negatives}.
\textbf{Relation-level false negative} (S3 in Figure~\ref{fig:re_example}) refers to the situation where there exists $t' = \langle s',r',o' \rangle \notin T_i$, but $r'$ is actually expressed by $\bci$, and does not appear in other $t\in T_i$.
Similarly, \textbf{Entity-level false negative} (S4 and S5 in Figure~\ref{fig:re_example}) means $r'$ appears in other $t \in T_i$.
% In other words, any missing triples will be considered relation-level false negative.
% TODO 如何解决关系标了 但是库缺失其他实体漏标。
% \textbf{FN} means that the negative labels used by entity extraction that are wrong.
% \textbf{Imbalanced data} in this paper specially refers to sample-level imbalance. That is, in most cases, $T_i$ only contains a small subset of $\mathcal{R}$.
\textbf{Imbalanced class distribution} means that the quantity of negative labels are much larger than positive ones.

\subsection{Addressing the False Negatives}
\label{sec:fn}
As shown in Table~\ref{tab:stat}, the triples in NYT (SKE) datasets\footnote{Detailed description of datasets is in Sec.~\ref{sec:data}} labeled by Freebase\footnote{\cite{bollacker2008freebase}}
(BaiduBaike\footnote{\url{https://baike.baidu.com/}}) is 88,253 (409,767), 
while the ones labeled by Wikidata\footnote{\cite{Vrandecic2014WikidataAF}} (CN-DBPedia\footnote{~\cite{Xu2017CNDBpediaAN}}) are 58,135 (342,931). 
In other words, there exists massive FN matches if only labeled by one KB due to the incompleteness of KBs. 
Notably, we find that the FN rate is underestimated by previous researches~\cite{min2013distant,xu2013filling}, based on the manual evaluation of which there are 15\%-35\% FN matches.
This discrepancy may be caused by human error.
In specific, a volunteer may accidentally miss some triples.
For example, as pointed out by Wei et al.~\shortcite[in Appendix C]{Wei2020ANC}, the test set of NYT11~\cite{Hoffmann2011KnowledgeBasedWS} missed lots of triples, especially when multiple relations occur in a same sentence, though labeled by human.
That also provides an evidence that FN's are harder to discover than FP's.

\begin{table}[!htbp]
  \centering
%   \addtolength{\tabcolsep}{-2.5pt}
  \small
        \begin{tabular}{lrrrr}
        \toprule
                 & \multicolumn{2}{c}{NYT (English)} & \multicolumn{2}{c}{SKE (Chinese)} \\
        \cmidrule{2-5}    \# Sentence & \multicolumn{2}{c}{56,196} & \multicolumn{2}{c}{194,747} \\
        \midrule
                 & \multicolumn{1}{l}{\# Triples} & \multicolumn{1}{l}{\# Rels} & \multicolumn{1}{l}{\# Triples} & \multicolumn{1}{l}{\# Rels} \\
        \cmidrule{2-5}    Original & 88,253   & 23       & 409,767  & 49 \\
        Re-labeled & 58,135   & 57       & 342,931  & 378 \\
        Intersection & 13,848   & 18       & 121,326  & 46 \\
        Union    & 132,540  & 62       & 631,372  & 381 \\
        \midrule
        Original FNR & $\ge$ 0.33     &          & $\ge$ 0.35     &  \\
        Relabel FNR & $\ge$ 0.56     &          & $\ge$ 0.46     &  \\
        \bottomrule
    \end{tabular}%
    \caption{Statistics of the quantity of distantly labeled relational triples by using different KB's. The ``original" refers to freebase for NYT and BaiduBaike for SKE. The ``relabeled" means aligning using Wikidata and CN-DBpedia to re-label NYT and SKE datasets.
    In specific, we consider triples with the same subject and object to be candidate triples and use a relation mapping table to determine whether the triples match.
    The intersection of SKE dataset has two values because the original relation has a one-to-many mapping with relations in CN-DBpedia. FNR stands for false negative rates, calculated by using the \# Triples in Original(Re-labeled) divided by the union.}
    % \addtolength{\tabcolsep}{2.5pt}
    \label{tab:stat}%
\end{table}%

% As a result, we observe that both the original source (i.e., freebase and BaiduBaike\footnote{\url{https://baike.baidu.com/}}, used to perform distant supervision by the original version of the two datasets.) and our newly applied source (i.e., Wikidata and CN-DBpedia) have labeled large quantity of triples.
% However, the intersection is only around 15\% - 35\% of the total number of triples aligned by each KB and the relation are not necessarilty the same.
% Such a small intersection means that each aligned dataset may result in either severe false negatives or potential data bias\footnote{By removing the ``NA" sentences}. 
% Besides, the relation varies between 
%1400 23个

% Table generated by Excel2LaTeX from sheet 'class-imbalance'
\begin{table*}[!htbp]
    \centering
    \small
      \begin{tabular}{lllrrrrrr}
          \toprule
          \multicolumn{1}{r}{\multirow{3}[6]{*}{Paradigm}} & \multicolumn{2}{c}{\multirow{2}[4]{*}{Theoretical}} & \multicolumn{2}{l}{NYT10-HRL} & \multicolumn{2}{l}{NYT11-HRL} & \multicolumn{2}{l}{SKE} \\
          \cmidrule{4-9}             & \multicolumn{2}{r}{} & \multicolumn{2}{l}{$|\mathcal{R}|$=31,$\bar{N}$= 39.08} & \multicolumn{2}{l}{$|\mathcal{R}|$=11, $\bar{N}$=39.46} & \multicolumn{2}{l}{$|\mathcal{R}|$=51,$\bar{N}$= 54.67} \\
          \cmidrule{2-9}             & $\pi_1$  & $\pi_2$  & \multicolumn{1}{l}{$\pi_1$} & \multicolumn{1}{l}{$\pi_2$} & \multicolumn{1}{l}{$\pi_1$} & \multicolumn{1}{l}{$\pi_2$} & \multicolumn{1}{l}{$\pi_1$} & \multicolumn{1}{l}{$\pi_2$} \\
          \midrule
          $s, o$  then $r$ & --       & $\ep[ \frac{\sum{y}}{|\mathcal{R}|}]$ & \multicolumn{1}{l}{--} & 0.01421  & \multicolumn{1}{l}{--} & 0.00280  & \multicolumn{1}{l}{--} & 0.00494 \\
          $s$  then $r,o$ & $\ep[ \frac{\sum{y}}{\bar{N}}]$ & $\ep[ \frac{\sum{y}}{\bar{N}*|\mathcal{R}|}]$ & 0.0585   & 0.00093  & 0.0574   & 0.00257  & 0.0405   & 0.00067 \\
          $r$ then $s,o$ & $\ep[ \frac{\sum{y}}{|\mathcal{R}|}]$ & $\ep[ \frac{\sum{y}}{4*\bar{N}}]$ & 0.0390   & 0.00842  & 0.0826   & 0.00835  & 0.0344   & 0.00927 \\
          \bottomrule
      \end{tabular}%
    \caption{Comparison of class prior under different relation extraction paradigms. $|\mathcal{R}|$ means the total number of relations and $\bar{N}$ is the average sentence length. $\pi_1$ ($\pi_2$) refers to the class prior for the first (second) task in the pipeline. $\pi_1$ for the first paradigm is omitted because it is often considered a preceding step. $\sum y$ is the summation of 1's in labels, of using which our intention is to represent the information a positive sample conveys.}
    \label{tab:neglabel}%
  \end{table*}%

\subsection{Addressing the Overwhelming Negative Labels}
\label{sec:neglabel}

We point out that some of the previous paradigms designed for relation extraction aggravate the imbalance and lead to inefficient supervision.
The mainstream approaches for relation extraction mainly fall into three paradigms depending on what to extract first.

\begin{itemize}
    \item[P1] The first paradigm is a pipeline that begins with named entity recognition (NER) and then classifies each entity pair into different relations, i.e., [$s,o$ then $r$]. 
    It is adopted by many traditional approaches~\cite{mintz2009distant,chan2011exploiting,zeng2014relation,Zeng2015DistantSF,gormley2015improved,dos2015classifying,lin2016neural}. 
    \item[P2] The second paradigm first detects all possible subjects in a sentence then identifies objects with respect to each relation, i.e., [$s$ then $r,o$]. Specific implementation includes modeling relation extraction as multi-turn question answering~\cite{Li2019EntityRelationEA}, span tagging~\cite{yu2020joint} and cascaded binary tagging \cite{Wei2020ANC}.
    % To note, the latter one is the current state-of-the-art.
    \item[P3] The third paradigm first perform sentence-level relation detection (cf. P1, which is at entity pair level.) then extract subjects and entities, i.e., [$r$ then $s,o$].
    This paradigm is largely unexplored.
    HRL~\cite{Takanobu2019AHF} is hitherto the only work to apply this paradigm based on our literature review.
\end{itemize}
We provide theoretical analysis of the output space and class prior with statistical support from three datasets (see Section~\ref{sec:data} for description) of the three paradigms in Table~\ref{tab:neglabel}.
The second step of P1 can be compared with the first step of P3.
Both of them find relation from a sentence (P1 with target entity pair given).
Suppose a sentence contains $m$ entities\footnote{Below the same.}, the classifier has to decide relation from $\mathcal{O}(m^2)$ entity pairs, while in reality, relations are often sparse, i.e., $\mathcal{O}(m)$. 
In other words, most entity pairs in P1 do not form valid relation, thus resulting in a low class prior.
The situation is even worse when the sentence contains more entities, such as in NYT11-HRL.
In addition, P1 is not sample efficient because the classifier will be trained/queried $\frac{m (m-1)}{2}$ for the same sentences.
For P2, we demonstrate with the problem formulation of \textsc{CasRel}~\cite{Wei2020ANC}.
The difference of the first-step class prior between P2 and P3 depends on the result of comparison between \# relations and average sentence length (i.e., $|\mathcal{R}|$ and $\bar{N}$), which varies in different scenarios/domains.
However, $\pi_2$ of P2 is extremely low, where a classifier has to decide from a space of $|\mathcal{R}|*\bar{N}$.
In contrast, P3 only has to decide from $4*\bar{N}$ based on our task formulation (Section~\ref{sec:pdef})

Other task formulations include jointly extracting the relation and entities \cite{yu2010jointly,Li2014IncrementalJE,miwa2014modeling,gupta2016table,katiyar2017going,Ren2017CoTypeJE} and recently in the manner of sequence tagging \cite{Zheng2017JointEO}, sequence-to-sequence learning \cite{Zeng2018ExtractingRF}.
In contrast to the aforementioned three paradigms, most of these methods actually provide an \textit{incomplete decision space} that cannot handle all the situation of relation extraction, for example, the overlapping one~\cite{Wei2020ANC}.
% Nevertheless, these approaches are not capable of handling overlapping relation extraction by design.

% Classification performed on this kind of imbalanced data is difficult.
% Fortunately, we show (in Section \ref{sec:pdef}) that the imbalanced data is caused by problem modeling, which is evitable by a carefully designed pipeline.
% Based on this idea, 

\section{Solution Framework}

\subsection{Framework of \mymethod}
\label{sec:pdef}

Given an instance $(c_i, T_i)$ from $\mathcal{D}$, the goal of training is to maximize the likelihood defined in Eq.~\eqref{eq:likelihood}. It is decomposed into two components by applying the definition of conditional probability, formulated in Eq. \eqref{eq:2stage}.

\begin{align}
\small
   \label{eq:likelihood}
   & \prod_{i=1}^{|\mathcal{D}|} \Pr(T_i|\bci;\theta) \\
  \label{eq:2stage}
  = & \prod_{i=1}^{|\mathcal{D}|} \prod_{r\in T_i} \Pr(r|\bci;\theta) \prod_{  \langle s,o \rangle \in T_i|r } \Pr(s,o|r,\bci;\theta),
\end{align}
where we use $r\in T_i$ as a shorthand for $r \in \{r \mid \sro\in T_i \}$, which means that $r$ occurs in the triple set w.r.t.\ $\bci$; Similarly, $s\in T_i$, $\langle s,o \rangle \in T_i|r$ stands for $s \in \{s \mid \sro\in T_i|r \}$ and $\langle s,o\rangle \in \{\langle s,o\rangle \mid \sro\in T_i|r \}$, respectively.
$T_i|r$ represents a subset of $T_i$ with a common relation $r$.
$\mathbbm{1}[\cdot]$ is an indicator function; $\mathbbm{1}[\text{condition}]=1$ when the condition happens.
We denote by $\theta$ the model parameters.

Under this decomposition, relational triple extraction task is formulate into two subtasks: \textit{relation classification} and \textit{entity extraction}.
\paragraph*{Relation Classification.} As is discussed, building relation classifier at entity-pair level will introduce excessive negative samples that forms a hard learning problem.
Therefore, we alternatively model the relation classification at sentence level.
Intuitively speaking, we hope that the model could capture \textit{what relation a sentence is expressing.}
We formalize it as a multi-label classification task. 
\begin{equation}
\label{eq:pr}
    \Pr(r|\bci;\theta) = \prod_{j=1}^{|\mathcal{R}|} (\hat{y}_{rc}^{j})^{\mathbbm{1}[{y_{rc}^j=1}]} (1-\hat{y}_{rc}^{j})^{\mathbbm{1}[{y_{rc}^j=0}]},
\end{equation}
where $\hat{y}_{rc}^{j}$ is the probability that $\mathbf{c}$ is expressing $r_j$, the $j$-th relation\footnote{$\hat{y}_{rc}^{j}$ is parameterized by $\theta$, omitted in the equation for clarity, below the same.}.
$y_{rc}^j$ is the ground truth from the labeled data; $y_{rc}^j = 1$ is equivalent to $r_j \in T_i$ while $y_{rc}^j = 0$ means the opposite.

\paragraph*{Entity Extraction.} 
We then model entity extraction task. 
We observe that given the relation $r$ and context $\bci$, it naturally forms a machine reading comprehension (MRC) task~\cite{chen2018neural}, where $(r, \bci, s/o)$ naturally fits into the paradigm of (\textsc{query, context, answer}).
Particularly, the subjects and objects are continuous spans from $\bci$, which falls into the category of \textit{span extraction}.
We adopt the boundary detection model with answer pointer~\cite{wang2016machine} as the output layer, which is widely used in MRC tasks.
Formally, for a sentence of $N$ tokens,
\begin{equation}
\label{eq:pso}
\begin{split}
    &\Pr(s,o|r,\bci;\theta)\\
    &=\prod_{k\in \mathcal{K}} \prod_{n=1}^{N} (\hat{y}_{ee}^{n,k})^{\mathbbm{1}[{y_{ee}^{n,k}=1}]} (1-\hat{y}_{ee}^{n,k})^{\mathbbm{1}[{y_{ee}^{n,k}=0}]},
\end{split}
\end{equation}
where $\mathcal{K}={\{s_{start},s_{end},o_{start},o_{end}\}}$ represents the identifier of each pointer; $\hat{y}_{ee}^{n,k}$ refers to the probability of $n$-th token being the start/end of the subject/object.
$y_{ee}^{n,k}$ is the ground truth from the training data;
if $\exists s \in T_i|r$ occurs in $\bci$ at position from $n$ to $n+l$, then $y_{ee}^{n,s_{start}} = 1$ and $y_{ee}^{n+l,s_{end}} = 1$, otherwise $0$; the same applies for the \textit{objects}.

\subsection{Advantages}
Our task formulation shows several advantages.
By adopting P3 as paradigm, the first and foremost advantage of our solution is that it suffers less from the imbalanced classes (Section~\ref{sec:neglabel}).
Secondly, relation-level false negative is easy to recover.
When modeled as a standard classification problem, many off-the-shelf methods on positive unlabeled learning can be leveraged.
Thirdly, entity-level false negatives do not affect relation classification.
Taking S5 in Figure~\ref{fig:re_example} as an example, even though the \textsc{Birthplace} relation between \textsc{William Swartz} and \textsc{Scranton} is missing, the relation classifier can still capture the signal from the other sample with a same relation, i.e., $\langle$ \textsc{Joe Biden, Birthplace, Scranton} $\rangle$.
Fourthly, this kind of modeling is easy to update with new relations without the need of retraining a model from bottom up.
Only relation classifier needs to be redesigned, while entity extractor can be updated in an online manner without modifying the model structure.
Last but not the least, relation classifier can be regarded as a pruning step when applied to practical tasks. Many existing methods treat relation extraction as question answering~\cite{Li2019EntityRelationEA,Zhao2020AskingEA}. 
However, without first identifying the relation, they all need to iterate over all the possible relations and ask diverse questions.
This results in extremely low efficiency where time consumed for predicting one sample may takes up to $|\mathcal{R}|$ times larger than our method.

% 好处：
% 显示建模FN问题。
% FN: Block relation-level false negative at RC model, 
% Unlike Wei \textit{et al.\} \cite{Wei2020ANC}, who model relation after  一个句子讲了多个关系，只要有一个被检测， 就是有的，合理减少FN （S5 例子）

% 新加关系不用重新写模型: 
% pruning: Many existing methods treat relation extraction as question answering\cite{Li2019EntityRelationEA,Zhao2020AskingEA}. However, without first identifying the relation, they all need to iterate over all the possible relations and ask diverse questions.
% 
% It naturally solves overlapping problem (including both SEO and EPO~\cite{}), though it is not the main selling point of this work.

% \subsection{Why does \mymethod Alleviate the False Negative?}

% We now compare the several different settings

\section{Our Model}
\label{sec:model}

\begin{figure*}[!ht]
  \centering
  \includegraphics[width=1.3\columnwidth]{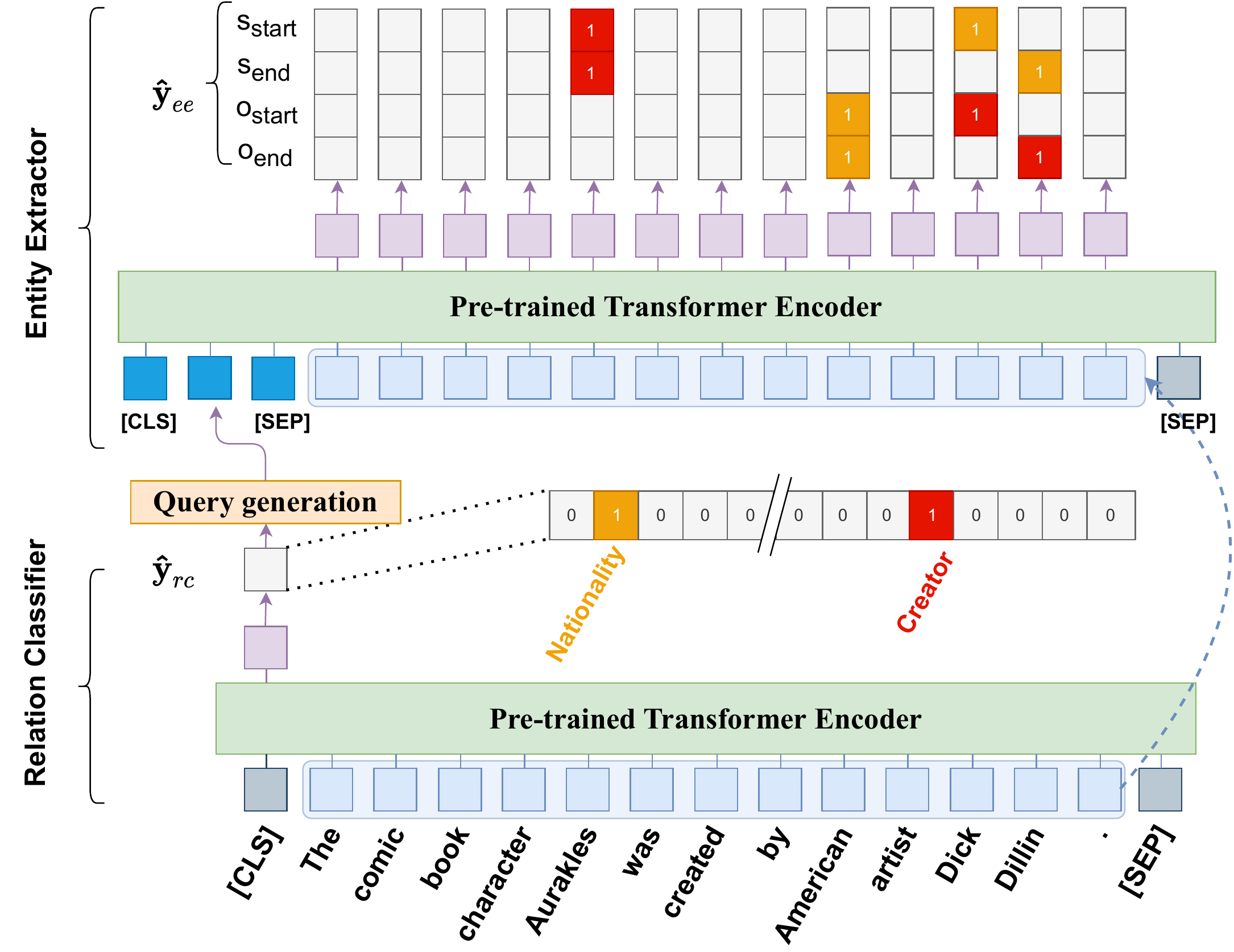}
  \caption{The overall architecture of \textsc{ReRe}. In this example, there are two relations, \textsc{Nationality}and \textsc{Creator}, can be found in the Relation Classifier, which will be sent to the Entity Extractor one by one along with the sentence. When The relation \textsc{Nationality} is extracted, the Entity Extractor will find the position of the subject and object of Nationality. The word \textsc{American} and \textsc{Dick Dillin} will be found. The relation \textsc{Creator} will then be handled similarly. The values of grey blocks in $\hyee$ are zero.}
  \label{fig:framework}
\end{figure*}

The relational triple extraction task decomposed in Eq.~\eqref{eq:2stage} inspires us to design a two-staged pipeline, in which we first detect relation at sentence level and then extract subjects/objects for each relation.
The overall architecture of \mymethod is shown in Figure \ref{fig:framework}.

% TODO 更新所有图

\subsection{Sentence-level Relation Classifier}
\label{sec:rc}
We first detect relation at sentence level.
The input is a sequence of tokens $\mathbf{c}$ and we denote by $\hyrc=[\hy_{rc}^1, \hy_{rc}^2,... , \hy_{rc}^{|\mathcal{R}|}]$ the output vector of the model, which aims to estimate $\hy_{rc}^i$ in Eq.~\eqref{eq:pr}.
We use BERT~\cite{Devlin2019BERTPO} for English and RoBERTa~\cite{liu2019roberta} for Chinese, pre-trained language models with multi-layer bidirectional Transformer structure \cite{vaswani2017attention}, to encode the inputs\footnote{For convenience, we refer to the pre-trained Transformer as BERT hereinafter.}.
Specifically, the input sequence $\mathbf{x}_{rc} = [ \text{\tt[CLS]}, \bci, \text{\tt[SEP]} ]$, which is fed into BERT for generating a token representation matrix $ \mathbf{H}_{rc} \in \mathbb{R}^{N \times d} $, where $d$ is the hidden dimension defined by pre-trained Transformers.
We take $\mathbf{h}_{rc}^0$, which is the encoded vector of the first token \texttt{[CLS]}, as the representation of the sentence.
The final output of relation classification module $\hyrc$ is defined in Eq.~\eqref{eq:yrc}.

\begin{equation}
\label{eq:yrc}
    \hyrc = \sigma( \bw{rc} \mathbf{h}_{rc}^0 + \bb{rc}  ),
\end{equation}
where $\bw{rc}$ and $\bb{rc}$ are trainable model parameters, representing weights and bias, respectively; $\sigma$ denotes the sigmoid activation function.

% Note that this module can be replaced by other model structure, such as CNN\cite{}, with or without pre-training.

% The truth labels are turned into an one hot vector $\yrc \in \{0,1\}^{|\mathcal{R}|}$.

\subsection{Relation-specific Entity Extractor}
After the relation detected at sentence-level, we extract subjects and objects for each candidate relation.
We aim to estimate $\hyee=[0,1]^{N\times4}$, of which each element corresponds to $\hat{y}_{ee}^{n,k}$ in Eq.~\eqref{eq:pso}, using a deep neural model.
% TODO 写MRC ，query,
We take $\hyrc$, the one-hot output vector of relation classifier, and generate query tokens $\mathbf{q}$ using each of the detected relations (i.e., the ``1''s in $\hyrc$). 
We are aware that many recent works~\cite{Li2019EntityRelationEA,Zhao2020AskingEA} have studied how to generate diverse queries for the given relation, which have the potential of achieving better performance.
Nevertheless, that is beyond the scope of this paper.
To keep things simple, we use the \textit{surface text} of a relation as the query.

Next, the input sequence is constructed as $\mathbf{x}_{ee} = [ \text{\tt[CLS]}, \mathbf{q}_i, \text{\tt[SEP]}, \mathbf{c}_i, \text{\tt[SEP]} ]$.
Like Section~\ref{sec:rc}, we get the token representation matrix $\mathbf{H}_{ee} \in \mathbb{R}^{N\times d}$ from BERT.
The $k$-th output pointer of entity extractor is defined by
\begin{equation}
\label{eq:yee}
    \hyee^k = \sigma( \mathbf{W}_{ee}^k \mathbf{H}_{ee} + \mathbf{b}_{ee}^k  ),
\end{equation}
where $k \in \{s_{start},s_{end},o_{start},o_{end}\}$ is in accordance to Eq.~\eqref{eq:pso}; $\bw{ee}^k$ and $\bb{ee}^k$ are the corresponding parameters.

The final subject/object spans are generated by pairing the nearest $s_{start}$/$o_{start}$ with $s_{end}$/$o_{end}$. 
Next, all subjects are paired to the nearest object. 
If multiple objects occur before the next subject appears, all subsequent objects will be paired with it until next subject occurs.
% During relation classification many False negative are recovered 

\subsection{Multi-label Collective Loss function}

In normal cases, the log-likelihood is taken as the learning objective.
However, as is emphasized, there exists many false negative samples in the training data.
Intuitively speaking, the negative labels cannot be simply considered as negative. 
Instead, a small portion of the negative labels should be considered as unlabeled positives and their influence towards the penalty should be eradicated.
Therefore, we adopt cPU~\cite{Xie2020CollectiveLF}, a collective loss function that is designed for \textit{positive unlabeled learning} (PU learning).
To briefly review, cPU considers the learning objective to be the
\textit{correctness} under a surrogate function,
\begin{equation}
\small
    \ell(\hy,y) = \ln(c(\hy,y)),
\end{equation}
where they redefine the \textit{correctness} function for PU learning as 
\begin{equation}
\small
    \label{eq:c}
    c(\hy,y) = \begin{cases}
        \ep[\hy] & \text{if } y=1,\\
        1-|\ep[\hy ] -\mu| & \text{otherwise},
    \end{cases}
\end{equation}
where $\mu$ is the ratio of false negative data (i.e., the unlabeled positive in the original paper).

We extend it to multi-label situation by embodying the original expectation at sample level. 
Due to the fact that class labels are highly imbalanced for our tasks, we introduce a class weight $\gamma \in (0,1)$ to downweight the positive penalty.
For relation classifier,
\begin{equation}
\small
    \ell_{rc}(\hby,\by) = \begin{cases}
    \displaystyle   
        -\gamma_{rc} \ln( \frac{1}{|\mathcal{R}|} \sum_{i=1}^{|\mathcal{R}|} \hat{y}_{rc}^i]) & \text{if } y_{rc}^i=1
    \\
    \displaystyle   
        - \ln( 1-|\frac{1}{|\mathcal{R}|}  \sum_{i=1}^{|\mathcal{R}|} \hat{y}_{rc}^i -\mu_{rc}|) & \text{otherwise}.
    \end{cases}
\end{equation}
For entity extractor,
\begin{equation}
\small
    \ell_{ee}(\hby^k,\by^k) = \begin{cases}
    \displaystyle
        - \gamma_{ee} \ln( \sum_{n=1}^{N} \hat{y}_{ee}^{n,k}]) & \text{if } y_{ee}^{n,k}=1
    \\
    \displaystyle
        -\ln( 1-|\sum_{n=1}^{N} \hat{y}_{ee}^{n,k} -\mu_{ee}|) & \text{otherwise}.
    \end{cases}
\end{equation}

In practice, we set $\mu=\pi(\tau + 1)$, where $\tau \approx 1-\frac{ \text{\# labeled positive} }{ \text{\# all positive} } $ is the ratio of false negative and $\pi$ is the class prior. 
% TODO list a table of data stats in appendix
Note that $\mu$ is not difficult to estimate for both relation classification and entity extraction task in practice.
Besides various of methods in the PU learning  \cite{Plessis2015ClasspriorEF,Bekker2018EstimatingTC} for estimating it, an easy approximation is $\mu \approx \pi$ when $\pi \ll \tau$, which happens to be the case for our tasks.

\section{Experiments}
\subsection{Datasets}
\label{sec:data}
Our experiments are conducted on these four datasets\footnote{We do not use WebNLG~\cite{Gardent2017CreatingTC} and ACE04\footnote{\url{https://catalog.ldc.upenn.edu/LDC2005T09}} because these datasets are not automatically labeled by distant supervision. WebNLG is constructed by natural language generation with triples. ACE04 is manually labeled.}.
Some statistics of the datasets are provided in Table~\ref{tab:stat} and Table~\ref{tab:neglabel}.
In relation extraction, some datasets with the same names involve different preprocessing, which leads to unfair comparison. 
We briefly review all the datasets below and specify the operations to perform before applying each dataset.
\begin{itemize}
    \normalsize
    \item \textbf{NYT} \cite{Riedel2010ModelingRA}. 
    NYT is the very first version among all the NYT-related datasets.
    It is based on the articles in New York Times\footnote{\url{https://www.nytimes.com/}}. 
    We use the sentences from it to conduct the pilot experiment in Table~\ref{tab:stat}.
    However, 1) it contains duplicate samples, e.g., 1504 in the training set; 2) It only labels the last word of an entity, which will mislead the evaluation results.
    \item \textbf{NYT10-HRL.} \& \textbf{NYT11-HRL.}
    These two datasets are based on NYT. 
    The difference is that they both contain \textit{complete} entity mentions.
    NYT10~\cite{Riedel2010ModelingRA} is the original one.
    and NYT11~\cite{Hoffmann2011KnowledgeBasedWS} is a small version of NYT10 with 53,395 training samples and a manually labeled test set of 368 samples.
    We refer to them as NYT10-HRL and NYT11-HRL after preprocessed by HRL~\cite{Takanobu2019AHF} where they removed 1) training relation not appearing in the testing and 2) ``NA" sentences.
    These two steps are almost adopted by all the compared methods.
    To compare fairly, we use this version in evaluations.

    \item \textbf{NYT21}. We provide relabel version of the test set of NYT11-HRL. 
    The test set of NYT11-HRL still have false negative problem.
    Most of the samples in the NYT11-HRL has only one relation. 
    We manually added back the missing triples to the test set.
    
    \item \textbf{SKE2019/SKE21}\footnote{\url{http://ai.baidu.com/broad/download?dataset=sked}}.
    SKE2019 is a dataset in Chinese published by Baidu.
    The reason we also adopt this dataset is that it is currently the largest dataset available for relation extraction.
    There are 194,747 sentences in the train set and 21,639 in the validate set.
    We manually labeled 1,150 sentences from the test set with 2,765 annotated triples, which we refer to as SKE21. 
    No preprocessing for this dataset is needed.
    We provide this data for future research\footnote{download url.}.
\end{itemize}

% Table generated by Excel2LaTeX from sheet 'exp_manual_test'
\begin{table*}[ht!]
  \centering
  \small
  \addtolength{\tabcolsep}{-2.5pt}
    \begin{tabular}{p{4.2cm}rrrrrrrrrlll}
    \toprule
             & \multicolumn{3}{c}{NYT10-HRL}  & \multicolumn{3}{c}{NYT11-HRL}  & \multicolumn{3}{c}{NYT21}      & \multicolumn{3}{c}{SKE21} \\
\cmidrule{2-13}             & \multicolumn{1}{c}{Prec.} & \multicolumn{1}{c}{Rec.} & \multicolumn{1}{c}{F1} & \multicolumn{1}{c}{Prec.} & \multicolumn{1}{c}{Rec.} & \multicolumn{1}{c}{F1} & \multicolumn{1}{c}{Prec.} & \multicolumn{1}{c}{Rec.} & \multicolumn{1}{c}{F1} & \multicolumn{1}{c}{Prec.} & \multicolumn{1}{c}{Rec.} & \multicolumn{1}{c}{F1} \\
    \midrule
    KB Match & 38.10       & 32.38    & 34.97    & 47.92    & 31.08    & 37.7     & 47.92    & 29.56    & 36.57    & 69.12    & 28.1     & 39.96 \\
    MultiR \cite{Hoffmann2011KnowledgeBasedWS} & \multicolumn{1}{c}{-} & \multicolumn{1}{c}{-} & \multicolumn{1}{c}{-} & 32.8     & 30.6     & 31.7     & \multicolumn{1}{c}{-} & \multicolumn{1}{c}{-} & \multicolumn{1}{c}{-} & \multicolumn{1}{c}{-} & \multicolumn{1}{c}{-} & \multicolumn{1}{c}{-} \\
    SPTree \cite{Miwa2016EndtoEndRE} & 49.2     & 55.7     & 52.2     & 52.2     & 54.1     & 53.1     & \multicolumn{1}{c}{-} & \multicolumn{1}{c}{-} & \multicolumn{1}{c}{-} & \multicolumn{1}{c}{-} & \multicolumn{1}{c}{-} & \multicolumn{1}{c}{-} \\
    NovelTagging \cite{Zheng2017JointEO} & 59.3     & 38.1     & 46.4     & 46.9     & 48.9     & 47.9     & \multicolumn{1}{c}{-} & \multicolumn{1}{c}{-} & \multicolumn{1}{c}{-} & \multicolumn{1}{c}{-} & \multicolumn{1}{c}{-} & \multicolumn{1}{c}{-} \\
    CoType \cite{Ren2017CoTypeJE} & \multicolumn{1}{c}{-} & \multicolumn{1}{c}{-} & \multicolumn{1}{c}{-} & 48.6     & 38.6     & 43.0       & \multicolumn{1}{c}{-} & \multicolumn{1}{c}{-} & \multicolumn{1}{c}{-} & \multicolumn{1}{c}{-} & \multicolumn{1}{c}{-} & \multicolumn{1}{c}{-} \\
    CopyR \cite{Zeng2018ExtractingRF} & 56.9     & 45.2     & 50.4     & 34.7     & 53.4     & 42.1     & \multicolumn{1}{c}{-} & \multicolumn{1}{c}{-} & \multicolumn{1}{c}{-} & \multicolumn{1}{c}{-} & \multicolumn{1}{c}{-} & \multicolumn{1}{c}{-} \\
    HRL \cite{Takanobu2019AHF} & 71.4     & 58.6     & 64.4     & 53.8     & 53.8     & 53.8     & \multicolumn{1}{c}{-} & \multicolumn{1}{c}{-} & \multicolumn{1}{c}{-} & \multicolumn{1}{c}{-} & \multicolumn{1}{c}{-} & \multicolumn{1}{c}{-} \\
    TPLinker \cite{wang2020tplinker}* & \textbf{81.19} & 65.41    & 72.45    & 56.2     & 55.14    & 55.67    & 59.78    & 55.78    & 57.71    & \multicolumn{1}{c}{-} & \multicolumn{1}{c}{-} & \multicolumn{1}{c}{-} \\
    CasRel \cite{Wei2020ANC}* & 77.7     & 68.8     & 73.0       & 50.1     & 58.4     & 53.9     & 58.64    & 56.62    & 57.61    & \multicolumn{1}{c}{-} & \multicolumn{1}{c}{-} & \multicolumn{1}{c}{-} \\
    \midrule
    \mymethod - LSTM  & 56.71    & 42.00       & 48.26    & \textbf{56.46} & 35.4     & 43.52    & \textbf{62.06} & 37.01    & 46.37    & \multicolumn{1}{c}{-} & \multicolumn{1}{c}{-} & \multicolumn{1}{c}{-} \\
    \mymethod & 75.45    & \textbf{72.50} & \textbf{73.95} & 53.12    & \textbf{59.59} & \textbf{56.23} & 57.69    & \textbf{61.69} & \textbf{59.62} & \multicolumn{1}{c}{\textbf{-}} & \multicolumn{1}{c}{\textbf{-}} & \multicolumn{1}{c}{\textbf{-}} \\
    \midrule
    \midrule
    TPLinker \cite{wang2020tplinker}*(exact) & \underline{80.34} & 65.11    & 71.93    & \underline{55.43} & 55.12    & 55.28    & \underline{58.96} & 55.78    & 57.33    & 83.86    & 84.77    & 84.32 \\
    CasRel \cite{Wei2020ANC}*(exact) & 75.12    & 65.72    & 70.11    & 47.88    & 55.13    & 51.25    & 55.06    & 54.49    & 54.78    & 86.94    & \underline{85.96} & 86.45 \\
    % \mymethod - LSTM (exact) & 53.99    & 39.99    & 45.95    & 52.59    & 32.97    & 40.53    & 57.33    & 34.19    & 42.83    & 69.9     & 53.65    & 60.71 \\
    \mymethod (exact) & 74.90     & \underline{71.97} & \underline{73.4} & 52.40     & \underline{58.91} & \underline{55.47} & 56.97    & \underline{60.93} & \underline{58.88} & \underline{90.44} & 84.20     & \underline{87.21} \\
    \bottomrule
    \end{tabular}%
\addtolength{\tabcolsep}{2.5pt}
  \caption{The main evaluation results of different models on NYT10-HRL,  NYT11-HRL, and two hand labeled test set NYT21 and SKE21 on by the compared method on the datasets. 
  The results with only one decimal are quoted from~\cite{Wei2020ANC}. The methods with * are based on our re-implementation.
  Best partial (exact) match results are marked \textbf{bold} (\underline{underlined}). }
  \label{tab:overall}%
\end{table*}%

\subsection{Compared Methods and Metrics}
We evaluate our model by comparing with several models on the same datasets, which are SOTA graphical model MultiR~\cite{Hoffmann2011KnowledgeBasedWS}, joint models  SPTree~\cite{Miwa2016EndtoEndRE} and NovelTagging~\cite{Zheng2017JointEO},
recent strong SOTA models CopyR~\cite{Zeng2018ExtractingRF}, HRL~\cite{Takanobu2019AHF}, CasRel~\cite{Wei2020ANC}, TPLinker~\cite{wang2020tplinker}.
We also provide the result of automatically aligning Wikidata/CN-KBpedia with the corpus, namely \textit{Match}, as a baseline.
To note, we only keep the intersected relations, otherwise it will result in low precision due to the false negative in the original dataset.
We report standard micro Precision (Prec.), Recall (Rec.) and F1 score for all the experiments.
Following the previous works~\cite{Takanobu2019AHF,Wei2020ANC}, we adopt partial match on these data sets for fair comparison. 
We also provide the results of exact match results of the methods we implemented, and only exact match on SKE2019.

\subsection{Overall Comparison}

We show the overall comparison result in Table~\ref{tab:overall}.
First, we observe that \mymethod consistently outperforms all the compared models.
We find an interesting result that by purely aligning the database with the corpus, it already achieves surprisingly good overall result (surpassing MultiR) and relatively high precision (comparable to CoType in NYT11-HRL).
However, the recall is quite low, which accords with our discussion in Section ~\ref{sec:fn} that distant supervision leads to many false negatives.
We also provide an ablation result where BERT is replaced with a bidirectional LSTM encoder~\cite{graves2013speech} with randomly initialized weights.
From the results we discover that even without BERT, our framework achieves competitive results against the previous approaches such as CoType and CopyR.
This further prove the effectiveness of our \mymethod  framework.

% DONE 更新实验数据, NYT-HRL11 用最好的。casrel算个exact match。

\begin{figure*}[!ht]
    \centering
    \begin{subfigure}
        \centering
        \includegraphics[width=0.99\columnwidth]{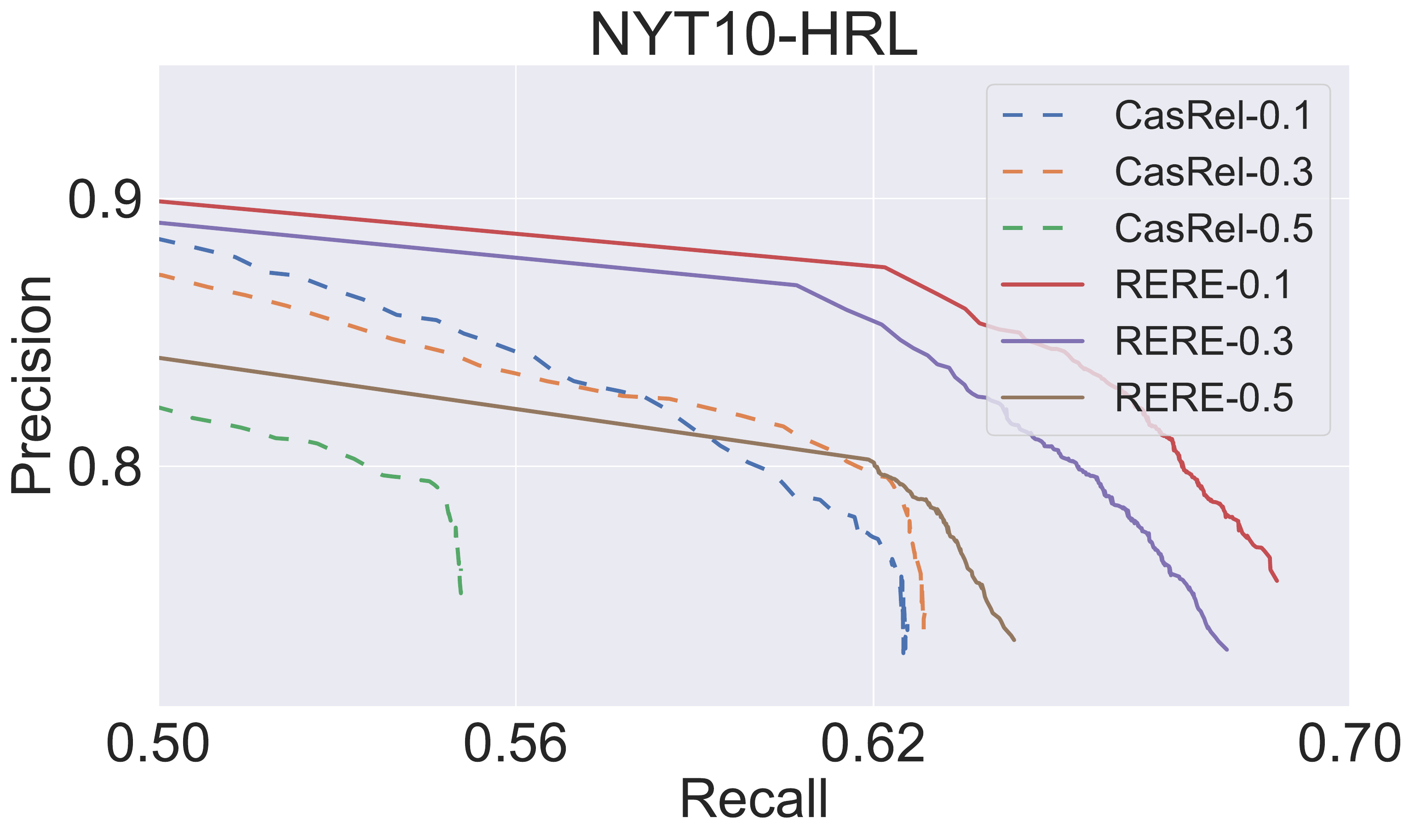}
    \end{subfigure}
    % \hfill
    \begin{subfigure}
        \centering
        \includegraphics[width=0.99\columnwidth]{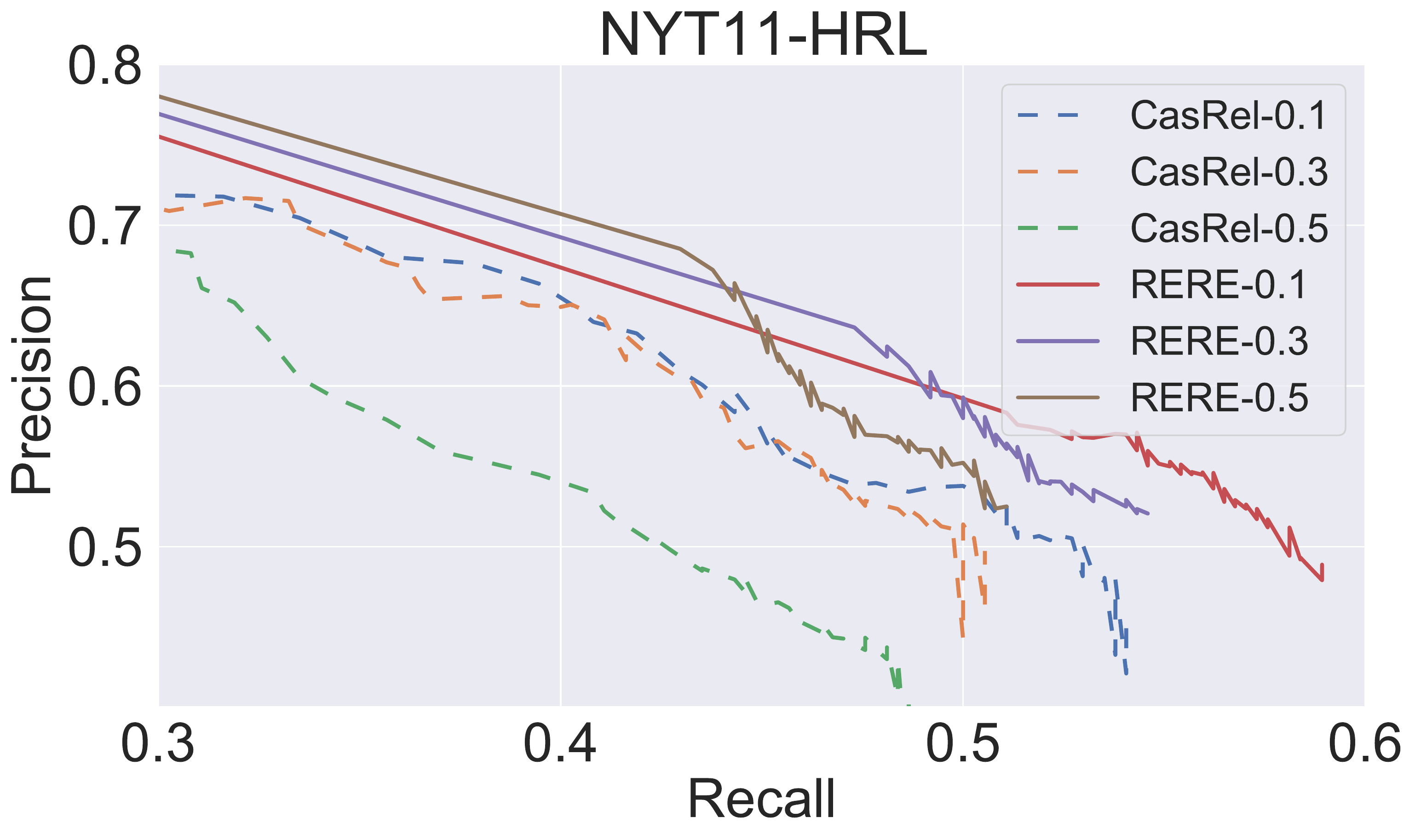}
    \end{subfigure}
    \caption{Precision-Recall Curve of \mymethod and \textsc{CasRel} under different false negative rate. 
    Lines are better in the upper-right corner than the opposite.
    Note that the coordinates do not starts from 0.}
    \label{fig:pr}
\end{figure*}

\subsection{How Robust is \mymethod against False Negatives?}

To further study how our model behaves when training data includes different quantity of false negatives, we conduct experiments on synthetic datasets.
We construct five new training data by randomly removing triples with probability of 0.1, 0.3 and 0.5, simulating the situation of different FN rates.
We show the precision-recall curves of our method in comparison with \textsc{CasRel}~\cite{Wei2020ANC}, the best performing competitor, in Figure~\ref{fig:pr}.
1) The overall performance of \mymethod is superior to competitor models even when trained on a dataset with a 0.5 FN rate.
2) We show that the intervals of \mymethod between lines are smaller than \textsc{CasRel}, indicating that the performance decline under different FN rates of \mymethod is smaller.
3) The straight line before curves of our model means that there is no data point at the places where recall is very low.
This means that our model is insensitive with the decision boundary and thus more robust. 

% \subsection{Ablation}

% TODO 
% Effectiveness of framework: withtout bert (用LSTM 替换)
% influence of $\mu_p$
% query(relation) generation
% 锁层参数

\section{Conclusion}
% \vspace{-4mm}
In this paper, we revisit the negative data in relation extraction task.
We first show that the false negative rate is largely underestimated by previous researches.
We then systematically compare three commonly adopted paradigms and prove that our paradigm suffers less from the overwhelming negative labels.
Based on this advantage, we propose \textsc{ReRe}, a pipelined framework that first detect relations at sentence level and then extract entities for each specific relation and provide a multi-label PU learning loss to recover false negatives.
Empirical results show that \mymethod consistently outperforms the existing state-of-the-arts by a considerable gap, even when learned with large false negative rates.

\bibliographystyle{acl_natbib}
\bibliography{refer}

\begin{thebibliography}{46}
\expandafter\ifx\csname natexlab\endcsname\relax\def\natexlab#1{#1}\fi

\bibitem[{Bekker and Davis(2018)}]{Bekker2018EstimatingTC}
Jessa Bekker and Jesse Davis. 2018.
\newblock Estimating the class prior in positive and unlabeled data through
  decision tree induction.
\newblock In \emph{Proceedings of AAAI}.

\bibitem[{Bollacker et~al.(2008)Bollacker, Evans, Paritosh, Sturge, and
  Taylor}]{bollacker2008freebase}
Kurt Bollacker, Colin Evans, Praveen Paritosh, Tim Sturge, and Jamie Taylor.
  2008.
\newblock Freebase: a collaboratively created graph database for structuring
  human knowledge.
\newblock In \emph{Proceedings of SIGMOD}.

\bibitem[{Chan and Roth(2011)}]{chan2011exploiting}
Yee~Seng Chan and Dan Roth. 2011.
\newblock Exploiting syntactico-semantic structures for relation extraction.
\newblock In \emph{Proceedings of ACL}, pages 551--560.

\bibitem[{Chen(2018)}]{chen2018neural}
Danqi Chen. 2018.
\newblock \emph{Neural Reading Comprehension and Beyond}.
\newblock Ph.D. thesis, Stanford University.

\bibitem[{Chowdhury and Lavelli(2012)}]{Chowdhury2012ImpactOL}
Md. Faisal~Mahbub Chowdhury and A.~Lavelli. 2012.
\newblock Impact of less skewed distributions on efficiency and effectiveness
  of biomedical relation extraction.
\newblock In \emph{Proceedings of COLING}.

\bibitem[{Devlin et~al.(2019)Devlin, Chang, Lee, and
  Toutanova}]{Devlin2019BERTPO}
J.~Devlin, Ming-Wei Chang, Kenton Lee, and Kristina Toutanova. 2019.
\newblock {BERT}: Pre-training of deep bidirectional transformers for language
  understanding.
\newblock In \emph{Proceedings of NAACL-HLT}.

\bibitem[{Feng et~al.(2018)Feng, Huang, Zhao, Yang, and
  Zhu}]{feng2018reinforcement}
Jun Feng, Minlie Huang, Li~Zhao, Yang Yang, and Xiaoyan Zhu. 2018.
\newblock Reinforcement learning for relation classification from noisy data.
\newblock In \emph{Proceedings of AAAI}, volume~32.

\bibitem[{Gardent et~al.(2017)Gardent, Shimorina, Narayan, and
  Perez-Beltrachini}]{Gardent2017CreatingTC}
Claire Gardent, Anastasia Shimorina, Shashi Narayan, and Laura
  Perez-Beltrachini. 2017.
\newblock Creating training corpora for {NLG} micro-planners.
\newblock In \emph{Proceedings of ACL}.

\bibitem[{Gormley et~al.(2015)Gormley, Yu, and Dredze}]{gormley2015improved}
Matthew~R Gormley, Mo~Yu, and Mark Dredze. 2015.
\newblock Improved relation extraction with feature-rich compositional
  embedding models.
\newblock In \emph{Proceedings of ACL}, pages 1774--1784.

\bibitem[{Graves et~al.(2013)Graves, Mohamed, and Hinton}]{graves2013speech}
Alex Graves, Abdel-rahman Mohamed, and Geoffrey Hinton. 2013.
\newblock Speech recognition with deep recurrent neural networks.
\newblock In \emph{2013 IEEE international conference on acoustics, speech and
  signal processing}, pages 6645--6649.

\bibitem[{Gupta et~al.(2016)Gupta, Sch{\"u}tze, and Andrassy}]{gupta2016table}
Pankaj Gupta, Hinrich Sch{\"u}tze, and Bernt Andrassy. 2016.
\newblock Table filling multi-task recurrent neural network for joint entity
  and relation extraction.
\newblock In \emph{Proceedings of COLING}, pages 2537--2547.

\bibitem[{Hoffmann et~al.(2011)Hoffmann, Zhang, Ling, Zettlemoyer, and
  Weld}]{Hoffmann2011KnowledgeBasedWS}
R.~Hoffmann, Congle Zhang, Xiao Ling, Luke Zettlemoyer, and Daniel~S. Weld.
  2011.
\newblock Knowledge-based weak supervision for information extraction of
  overlapping relations.
\newblock In \emph{Proceedings of ACL}.

\bibitem[{Jia et~al.(2019)Jia, Dai, Xiao, and Wu}]{jia2019arnor}
Wei Jia, Dai Dai, Xinyan Xiao, and Hua Wu. 2019.
\newblock {ARNOR}: Attention regularization based noise reduction for distant
  supervision relation classification.
\newblock In \emph{Proceedings of ACL}, pages 1399--1408.

\bibitem[{Katiyar and Cardie(2017)}]{katiyar2017going}
Arzoo Katiyar and Claire Cardie. 2017.
\newblock Going out on a limb: Joint extraction of entity mentions and
  relations without dependency trees.
\newblock In \emph{Proceedings of ACL}, pages 917--928.

\bibitem[{Li and Ji(2014)}]{Li2014IncrementalJE}
Q.~Li and Heng Ji. 2014.
\newblock Incremental joint extraction of entity mentions and relations.
\newblock In \emph{Proceedings of ACL}.

\bibitem[{Li et~al.(2020)Li, Sun, Meng, Liang, Wu, and Li}]{Li2020DiceLF}
Xiaoya Li, Xiaofei Sun, Yuxian Meng, Junjun Liang, F.~Wu, and J.~Li. 2020.
\newblock Dice loss for data-imbalanced nlp tasks.
\newblock In \emph{Proceedings of ACL}.

\bibitem[{Li et~al.(2019)Li, Yin, Sun, Li, Yuan, Chai, Zhou, and
  Li}]{Li2019EntityRelationEA}
Xiaoya Li, Fan Yin, Zijun Sun, Xiayu Li, Arianna Yuan, Duo Chai, Mingxin Zhou,
  and J.~Li. 2019.
\newblock Entity-relation extraction as multi-turn question answering.
\newblock In \emph{Proceedings of ACL}.

\bibitem[{Lin et~al.(2018)Lin, Lu, Han, and Sun}]{lin2018adaptive}
Hongyu Lin, Yaojie Lu, Xianpei Han, and Le~Sun. 2018.
\newblock Adaptive scaling for sparse detection in information extraction.
\newblock In \emph{Proceedings of ACL}, pages 1033--1043.

\bibitem[{Lin et~al.(2016)Lin, Shen, Liu, Luan, and Sun}]{lin2016neural}
Yankai Lin, Shiqi Shen, Zhiyuan Liu, Huanbo Luan, and Maosong Sun. 2016.
\newblock Neural relation extraction with selective attention over instances.
\newblock In \emph{Proceedings of ACL}, pages 2124--2133.

\bibitem[{Liu et~al.(2019)Liu, Ott, Goyal, Du, Joshi, Chen, Levy, Lewis,
  Zettlemoyer, and Stoyanov}]{liu2019roberta}
Yinhan Liu, Myle Ott, Naman Goyal, Jingfei Du, Mandar Joshi, Danqi Chen, Omer
  Levy, Mike Lewis, Luke Zettlemoyer, and Veselin Stoyanov. 2019.
\newblock Roberta: A robustly optimized bert pretraining approach.
\newblock \emph{arXiv preprint arXiv:1907.11692}.

\bibitem[{Min et~al.(2013)Min, Grishman, Wan, Wang, and
  Gondek}]{min2013distant}
Bonan Min, Ralph Grishman, Li~Wan, Chang Wang, and David Gondek. 2013.
\newblock Distant supervision for relation extraction with an incomplete
  knowledge base.
\newblock In \emph{Proceedings of HLT-NAACL}.

\bibitem[{Mintz et~al.(2009)Mintz, Bills, Snow, and
  Jurafsky}]{mintz2009distant}
Mike Mintz, Steven Bills, Rion Snow, and Dan Jurafsky. 2009.
\newblock Distant supervision for relation extraction without labeled data.
\newblock In \emph{Proceedings of ACL}.

\bibitem[{Miwa and Bansal(2016)}]{Miwa2016EndtoEndRE}
Makoto Miwa and Mohit Bansal. 2016.
\newblock End-to-end relation extraction using lstms on sequences and tree
  structures.
\newblock In \emph{Proceedings of ACL}, pages 1105--1116.

\bibitem[{Miwa and Sasaki(2014)}]{miwa2014modeling}
Makoto Miwa and Yutaka Sasaki. 2014.
\newblock Modeling joint entity and relation extraction with table
  representation.
\newblock In \emph{Proceedings of EMNLP}, pages 1858--1869.

\bibitem[{du~Plessis et~al.(2015)du~Plessis, Niu, and
  Sugiyama}]{Plessis2015ClasspriorEF}
Marthinus~Christoffel du~Plessis, Gang Niu, and Masashi Sugiyama. 2015.
\newblock Class-prior estimation for learning from positive and unlabeled data.
\newblock \emph{Machine Learning}, 106:463--492.

\bibitem[{Ren et~al.(2017)Ren, Wu, He, Qu, Voss, Ji, Abdelzaher, and
  Han}]{Ren2017CoTypeJE}
Xiang Ren, Zeqiu Wu, Wenqi He, Meng Qu, Clare~R Voss, Heng Ji, Tarek~F
  Abdelzaher, and Jiawei Han. 2017.
\newblock Cotype: Joint extraction of typed entities and relations with
  knowledge bases.
\newblock In \emph{Proceedings of WWW}, pages 1015--1024.

\bibitem[{Riedel et~al.(2010)Riedel, Yao, and McCallum}]{Riedel2010ModelingRA}
S.~Riedel, Limin Yao, and A.~McCallum. 2010.
\newblock Modeling relations and their mentions without labeled text.
\newblock In \emph{Proceedings of ECML/PKDD}.

\bibitem[{Roller et~al.(2015)Roller, Agirre, Soroa, and
  Stevenson}]{roller2015improving}
Roland Roller, Eneko Agirre, Aitor Soroa, and Mark Stevenson. 2015.
\newblock Improving distant supervision using inference learning.
\newblock In \emph{Proceedings of ACL}, pages 273--278.

\bibitem[{dos Santos et~al.(2015)dos Santos, Xiang, and
  Zhou}]{dos2015classifying}
Cicero dos Santos, Bing Xiang, and Bowen Zhou. 2015.
\newblock Classifying relations by ranking with convolutional neural networks.
\newblock In \emph{Proceedings of ACL}, pages 626--634.

\bibitem[{Takanobu et~al.(2019)Takanobu, Zhang, Liu, and
  Huang}]{Takanobu2019AHF}
Ryuichi Takanobu, Tianyang Zhang, Jiexi Liu, and Minlie Huang. 2019.
\newblock A hierarchical framework for relation extraction with reinforcement
  learning.
\newblock In \emph{Proceedings of AAAI}, volume~33, pages 7072--7079.

\bibitem[{Vaswani et~al.(2017)Vaswani, Shazeer, Parmar, Uszkoreit, Jones,
  Gomez, Kaiser, and Polosukhin}]{vaswani2017attention}
Ashish Vaswani, Noam Shazeer, Niki Parmar, Jakob Uszkoreit, Llion Jones,
  Aidan~N Gomez, {\L}ukasz Kaiser, and Illia Polosukhin. 2017.
\newblock Attention is all you need.
\newblock In \emph{Proceedings of NeuroIPS}, pages 6000--6010.

\bibitem[{Vrandecic and Kr{\"o}tzsch(2014)}]{Vrandecic2014WikidataAF}
Denny Vrandecic and M.~Kr{\"o}tzsch. 2014.
\newblock Wikidata: a free collaborative knowledgebase.
\newblock \emph{Communications of the ACM}, 57:78--85.

\bibitem[{Wang and Jiang(2017)}]{wang2016machine}
Shuohang Wang and Jing Jiang. 2017.
\newblock Machine comprehension using match-lstm and answer pointer.
\newblock In \emph{Proceedings of ICLR}.

\bibitem[{Wang et~al.(2020)Wang, Yu, Zhang, Liu, Zhu, and
  Sun}]{wang2020tplinker}
Yucheng Wang, Bowen Yu, Yueyang Zhang, Tingwen Liu, Hongsong Zhu, and Limin
  Sun. 2020.
\newblock {TPLinker}: Single-stage joint extraction of entities and relations
  through token pair linking.
\newblock In \emph{Proceedings of COLING}, pages 1572--1582.

\bibitem[{Wei et~al.(2020)Wei, Su, Wang, Tian, and Chang}]{Wei2020ANC}
Zhepei Wei, Jianlin Su, Yue Wang, Y.~Tian, and Yi~Chang. 2020.
\newblock A novel cascade binary tagging framework for relational triple
  extraction.
\newblock In \emph{Proceedings of ACL}.

\bibitem[{Xie et~al.(2020)Xie, Cheng, Liang, Chen, and
  Xiao}]{Xie2020CollectiveLF}
Chenhao Xie, Qiao Cheng, Jiaqing Liang, Lihan Chen, and Y.~Xiao. 2020.
\newblock Collective loss function for positive and unlabeled learning.
\newblock \emph{ArXiv}, abs/2005.03228.

\bibitem[{Xu et~al.(2017)Xu, Xu, Liang, Xie, Liang, Cui, and
  Xiao}]{Xu2017CNDBpediaAN}
Bo~Xu, Yong Xu, Jiaqing Liang, Chenhao Xie, Bin Liang, Wanyun Cui, and Y.~Xiao.
  2017.
\newblock {CN-DBpedia}: A never-ending chinese knowledge extraction system.
\newblock In \emph{Proceedings of IEA/AIE}.

\bibitem[{Xu et~al.(2013)Xu, Hoffmann, Zhao, and Grishman}]{xu2013filling}
Wei Xu, Raphael Hoffmann, Le~Zhao, and Ralph Grishman. 2013.
\newblock Filling knowledge base gaps for distant supervision of relation
  extraction.
\newblock In \emph{Proceedings of ACL}, pages 665--670.

\bibitem[{Yu et~al.(2020)Yu, Zhang, Shu, Liu, Wang, Wang, and Li}]{yu2020joint}
Bowen Yu, Zhenyu Zhang, Xiaobo Shu, Tingwen Liu, Yubin Wang, Bin Wang, and
  Sujian Li. 2020.
\newblock Joint extraction of entities and relations based on a novel
  decomposition strategy.
\newblock In \emph{Proceedings of ECAI}.

\bibitem[{Yu and Lam(2010)}]{yu2010jointly}
Xiaofeng Yu and Wai Lam. 2010.
\newblock Jointly identifying entities and extracting relations in encyclopedia
  text via a graphical model approach.
\newblock In \emph{Proceedings of COLING}, pages 1399--1407.

\bibitem[{Zeng et~al.(2015)Zeng, Liu, Chen, and Zhao}]{Zeng2015DistantSF}
Daojian Zeng, Kang Liu, Yubo Chen, and Jun Zhao. 2015.
\newblock Distant supervision for relation extraction via piecewise
  convolutional neural networks.
\newblock In \emph{Proceedings of EMNLP}.

\bibitem[{Zeng et~al.(2014)Zeng, Liu, Lai, Zhou, Zhao
  et~al.}]{zeng2014relation}
Daojian Zeng, Kang Liu, Siwei Lai, Guangyou Zhou, Jun Zhao, et~al. 2014.
\newblock Relation classification via convolutional deep neural network.
\newblock In \emph{Proceedings of COLING}, pages 2335--2344.

\bibitem[{Zeng et~al.(2018{\natexlab{a}})Zeng, He, Liu, and
  Zhao}]{zeng2018large}
Xiangrong Zeng, Shizhu He, Kang Liu, and Jun Zhao. 2018{\natexlab{a}}.
\newblock Large scaled relation extraction with reinforcement learning.
\newblock In \emph{Proceedings of AAAI}, volume~32.

\bibitem[{Zeng et~al.(2018{\natexlab{b}})Zeng, Zeng, He, Liu, and
  Zhao}]{Zeng2018ExtractingRF}
Xiangrong Zeng, Daojian Zeng, Shizhu He, Kang Liu, and Jun Zhao.
  2018{\natexlab{b}}.
\newblock Extracting relational facts by an end-to-end neural model with copy
  mechanism.
\newblock In \emph{Proceedings of ACL}.

\bibitem[{Zhao et~al.(2020)Zhao, Yan, Cao, and Li}]{Zhao2020AskingEA}
Tianyang Zhao, Zhao Yan, Y.~Cao, and Zhoujun Li. 2020.
\newblock Asking effective and diverse questions: A machine reading
  comprehension based framework for joint entity-relation extraction.
\newblock In \emph{Proceedings of IJCAI}.

\bibitem[{Zheng et~al.(2017)Zheng, Wang, Bao, Hao, Zhou, and
  Xu}]{Zheng2017JointEO}
Suncong Zheng, Feng Wang, Hongyun Bao, Yuexing Hao, Peng Zhou, and Bo~Xu. 2017.
\newblock Joint extraction of entities and relations based on a novel tagging
  scheme.
\newblock In \emph{Proceedings of ACL}, pages 1227--1236.

\end{thebibliography}

%\appendix

\end{document}